\documentclass[conference]{IEEEtran}
\IEEEoverridecommandlockouts
\usepackage{cite}
\usepackage{amsmath,amssymb,amsfonts}
\usepackage{algorithmic}
\usepackage{graphicx}
\usepackage{textcomp}
\usepackage{xcolor}
\usepackage{balance}
\usepackage{csquotes}
\usepackage{soul}
\usepackage{url}
\usepackage{tabularx}
\usepackage{booktabs} 
\usepackage{enumitem}
\usepackage{fontawesome}
\usepackage{array} 
\usepackage{multirow} 
\usepackage{amsmath}
\usepackage{amssymb}
\usepackage{hyperref}

\newcolumntype{P}[1]{>{\centering\arraybackslash}p{#1}}

\def\BibTeX{{\rm B\kern-.05em{\sc i\kern-.025em b}\kern-.08em
    T\kern-.1667em\lower.7ex\hbox{E}\kern-.125emX}}

\def\eg{\emph{e.g., }} 
\def\ie{\emph{i.e., }}

\newcommand{\pquotes}[1]{\textcolor[gray]{0.35}{\textit{#1}}}

\makeatletter
\newcommand{\linebreakand}{%
  \end{@IEEEauthorhalign}
  \hfill\mbox{}\par
  \mbox{}\hfill\begin{@IEEEauthorhalign}
}
\makeatother

\begin{document}

\title{The Design of On-Body Robots for Older Adults
\thanks{
\textbf{Authorship Statement.} Conceptualization, Methodology, Validation, Writing - Review \& Editing (all authors), Investigation (VNA, CJ, JL), Formal analysis, Data curation (VNA, CJ), Visualization (VNA, CJ, CH), Writing-Original Draft, Project Administration (VNA), Resources (CH, HP), Supervision (CH, AO, HP, GG), Funding (CH). \textbf{AI Use.} Text edited with LLM; output checked for correctness by authors.}
}



\author{\IEEEauthorblockN{Victor Nikhil Antony}
\IEEEauthorblockA{
\textit{Johns Hopkins University}\\
Baltimore, MD, USA \\
vantony1@jhu.edu}
\and
\IEEEauthorblockN{Clara Jeon}
\IEEEauthorblockA{
\textit{Johns Hopkins University}\\
Baltimore, MD, USA \\
cjeon6@jh.edu}
\and
\IEEEauthorblockN{Jiasheng Li}
\IEEEauthorblockA{
\textit{University of Maryland}\\
College Park, MD, USA \\
jsli@umd.edu}
\and
\IEEEauthorblockN{Ge Gao}
\IEEEauthorblockA{
\textit{University of Maryland}\\
College Park, MD, USA \\
gegao@umd.edu}
\linebreakand
\IEEEauthorblockN{Huaishu Peng}
\IEEEauthorblockA{
\textit{University of Maryland}\\
College Park, MD, USA \\
huaishu@umd.edu}
\and
\IEEEauthorblockN{Anastasia K. Ostrowski}
\IEEEauthorblockA{
\textit{Purdue University}\\
West Lafayette, IN, USA \\
akostrow@purdue.edu}
\and
\IEEEauthorblockN{Chien-Ming Huang}
\IEEEauthorblockA{
\textit{Johns Hopkins University}\\
Baltimore, MD, USA \\
chienming.huang@jhu.edu}
}


\maketitle


\begin{abstract}
Wearable technology has significantly improved the quality of life for older adults, and the emergence of on-body, movable robots presents new opportunities to further enhance well-being. Yet, the interaction design for these robots remains under-explored, particularly from the perspective of older adults. We present findings from a two-phase co-design process involving 13 older adults to uncover design principles for on-body robots for this population. We identify a rich spectrum of potential applications and characterize a design space to inform how on-body robots should be built for older adults. Our findings highlight the importance of considering factors like co-presence, embodiment, and multi-modal communication. Our work offers design insights to facilitate the integration of on-body robots into daily life and underscores the value of involving older adults in the co-design process to promote usability and acceptance of emerging wearable robotic technologies.
\end{abstract}


\section{Introduction}

Wearable technology, such as smartwatches, glucose monitors, and hearing aids, has played a vital role in improving the quality of life for older adults by offering benefits like enhancing socio-emotional and cognitive functions, reducing depression, and promoting self-awareness and behavior change \cite{mulrow1990quality, litchman2017real, chung2023community}. Emerging wearables like on-skin sensors \cite{kim2011epidermalelectronics, weigel2015iskin}, implanted interfaces \cite{kao2015nailo, holz2012implanted}, and digital fabrics \cite{poupyrev2016project} are unlocking new capabilities in sensing, interaction, and expression \cite{vega2014beauty}. As wearable technology continues to miniaturize and integrate closer with the human body, wearables are poised to play a pivotal role in enabling graceful aging \cite{orlov2012technology, fournier2020designing}.

Wearable robotic systems with their proximity to the user's body can enable new interaction paradigms and unique opportunities for empowering older adults. To date, wearable robots for older adults have primarily been in the form of exoskeletons, which are designed to enhance mobility and improve gait \cite{kong2006design, jayaraman2022modular, jung2019older}. A distinct class of wearable robots—referred to in this work as \textit{on-body robots}—differs from exoskeletons in two key aspects: their compact form factor, and their ability to move around the body \cite{sathya2022calico, dementyev2017skinbot, dementyev2016rovables}. On-Body, movable robots such as \textit{Calico}\cite{sathya2022calico}, \textit{SkinBot}\cite{weigel2015iskin}, and \textit{Rovables}\cite{dementyev2016rovables} have existed for several years, yet many open questions remain regarding their interaction design for the aging populations: What roles could these robots play in older adults' lives? How would they communicate effectively with users? How can their presence be made comfortable and seamless?

\begin{figure}
    \centering
    \includegraphics[width=\linewidth]{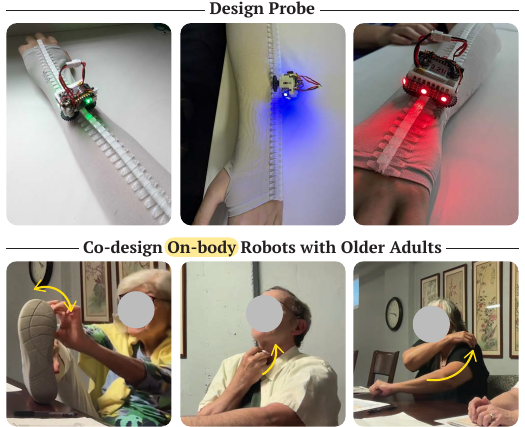}
    \caption{We engaged older adults as co-designers of on-body robots using \textit{Calico}\cite{sathya2022calico} as a design probe to ground the design process.}
    \label{fig:teaser}
    \vspace{-1em}
\end{figure}

Proximity to the human body signifies a deep sense of comfort and trust, typically reserved for only the most essential and intimate entities \cite{hall1990hidden}. The ability of on-body robots to move across the body, combined with their physical closeness, presents unique opportunities to empower older adults by building on the benefits of existing wearable technologies. While this closeness offers potential for fostering rich human-robot relationships, it also demands a high level of trust and reliance for successful adoption. Moreover, this proximity requires interactions so seamlessly integrated into the human experience they are perceived as an extension of the body. For on-body robots to be successfully integrated into the lives of older adults, the design of on-body interactions must be grounded in principles that reflect the specific needs and preferences of this population. Leveraging co-design as a methodology can uncover these principles by actively involving stakeholders (\ie older adults) in the design process \cite{muller1993participatory, rogers2022maximizing, antony2023codesign}, enhancing the usability and adoption of on-body robots.

In this work, we engage older adults as designers of on-body robots through a dual-phase co-design process. First, we explore the use cases and design of on-body robots broadly (\textit{divergence}), followed by application-focused design workshops to understand the finer-grained interaction needs of these systems (\textit{convergence}). Through our co-design work, we make the following contributions:
\begin{enumerate}[leftmargin=*]
    \item An initial framework of the design space for on-body robots based on design insights gathered from and interactions designed by older adults to guide future research.
    \item Reflections on involving older adults in designing on-body robots, drawing from our design process learnings.
\end{enumerate}


\section{Related Works}

\subsection{Wearable Robots}

Wearable robots vary widely in form and function. Exoskeletons, a prominent class, focus on mobility assistance and rehabilitation, targeting specific areas such as shoulders or hands \cite{oneill2017exoshoulder, gao2023portable, in2015exo}, or providing broader support to regions such as the lower limbs \cite{huo2014exolower, asbeck2014stronger}. Other wearable robots offer physical augmentation, such as a third thumb \cite{zhou2019novel, shafti2021playing} or arm \cite{muehlhaus2023need, vatsal2017wearing}, delivering active assistance while remaining fixed in place. Moreover, stationary robots perched on the shoulder have been studied as wearable companions \cite{tsumaki201220, jiang2019survey}.


A newer class of wearable robots consists of on-body, locomotive systems, capable of moving across the body, offering greater flexibility in interaction. These robots can move via direct skin contact \cite{dementyev2017skinbot}, climb on clothing \cite{dementyev2016rovables, birkmeyer2011clash}, or travel along tracks embedded in garments \cite{sathya2022calico}. This mobility sets them apart by enabling dynamic, on-body interactions and unlocking new possibilities for user engagement. While the design needs of exoskeletons have been explored \cite{jung2019older}, effective interaction paradigms for movable, on-body robots remains unexplored. Developing these paradigms is essential if on-body robots would be developed to support older adults.

\subsection{Designing with Older Adults}

Co-design is an effective methodology for understanding the design needs of special populations, such as older adults, by leveraging their lived experiences \cite{rogers2022maximizing, alves2015social, ostrowski2021personal}. This flexible approach allows stakeholders to act as users, testers, informants, partners, or co-researchers \cite{alves2021children, nanavati2023design}. Its versatility spans contexts ranging from assistive robots for aging in place \cite{lee2018reframing, alves2015social} and robots for dementia or depression \cite{moharana2019robots, lee2017steps}, to challenges like promoting physical activity \cite{antony2023codesign}, designing fitness apps \cite{harrington2018designing}, and improving wearable tech adoption \cite{nevay2015role}.

The flexibility of co-design extends to the diverse range of activities it supports, such as sketching \cite{lee2017steps}, story-boarding \cite{bjorling2019participatory}, mind-mapping \cite{antony2023codesign}, prototyping \cite{bjorling2019participatory, lee2017steps}, worksheets \cite{axelsson2021social} and role-playing \cite{bjorling2019participatory}. This adaptability allows for the selection of design exercises that foster both divergent and convergent design thinking \cite{ostrowski2021long}. In addition to design exercises, low-fidelity design probes can further inspire design thinking and offer new insights for future technologies \cite{hutchinson2003technology, gough2021co}. To effectively engage older adults as designers of on-body robots, we employ a co-design approach that incorporates various design activities to foster both convergent and divergent thinking, with a design probe used to anchor the design process.


\section{Design Process}
\begin{figure*}[h]
\centering
\includegraphics[width=\textwidth]{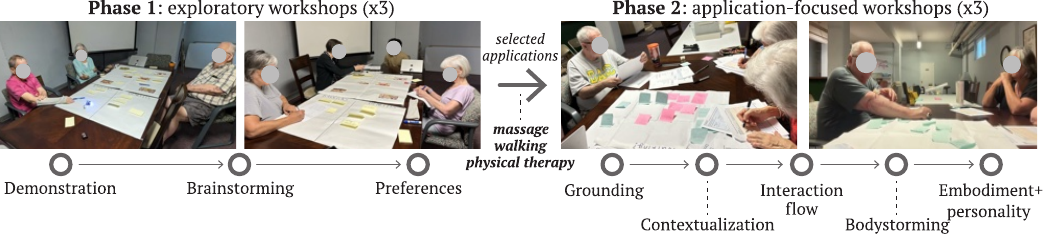}
\caption{Our two-phased design process consisted of three exploratory workshops for divergent ideation of use cases for on-body robots and application-focused workshops for convergent, detailed interaction design for three domains: \textit{massage, physical therapy} and \textit{walking}.}
\label{fig:process}
\end{figure*}

We employed a two-phase co-design process to explore the design space of on-body robots. The first phase involved exploratory workshops that encouraged open-ended, broad exploration of on-body robots (\textit{divergence}). The second phase consisted of workshops, where participants engaged in application-driven design exercises (\textit{convergence}) (see Fig. \ref{fig:process}).

\textbf{Design Probe.} We used \textit{Calico} \cite{sathya2022calico}, a small robot that moves around the body on flexible 3D-printed tracks embedded in clothing and communicate via LEDs strips, as our design probe. Calico’s simple setup and Wizard-of-Oz interface enabled us to introduce the on-body robot concept to participants, providing a tangible foundation for design \cite{ostrowski2021long}.

\textbf{Participants.} We recruited 13 independently living older adults to engage in the co-design of on-body robots. Each participant is assigned a pseudonym instead of depersonalized IDs (\eg p1) \cite{lee2018reframing} (see Table \ref{tab:participants_short}). The workshops were approved by our institutional review board and compensated at 15USD/hr. The sole inclusion criterion was age of 65 or older.

\begin{table}[h]
\centering
\caption{Overview of Older Adult Participant Demographics}
\label{tab:participants_short}
\resizebox{\columnwidth}{!}{%
\begin{tabular}{p{1.0cm}|p{0.8cm}|p{0.5cm}|p{1.2cm}|p{1.65cm}|p{1.8cm}} 
    \hline
    \textbf{Pseudo} & \textbf{Gender} & \textbf{Age} & \textbf{Ethnicity} & \textbf{Lives} & \textbf{Workshops} \\ 
    \hline
    \hline
    Rachel     & Female & 66 & Caucasian & Alone & EW \#2, AW \#2\\ 
    \hline
    Sylvia     & Female & 70 & Caucasian & with Partner & AW \#2\\ 
    \hline
    Leona      & Female & 81 & Caucasian & Alone & EW \#1, AW \#3\\ 
    \hline
    Raymond    & Male & 76 & Caucasian & Alone & EW \#1, AW \#2\\ 
    \hline
    Melanie    & Female & 76 & Caucasian & Alone & AW \#3\\ 
    \hline
    Martin     & Male & 75 & Caucasian & with Partner & AW \#3\\ 
    \hline
    Randall    & Male & 73 & Caucasian & with Partner & EW \#1, AW \#1\\ 
    \hline
    Norman     & Male & 69 & Caucasian & with Partner & AW \#1\\ 
    \hline
    Camille    & Male & 76  & Caucasian & Alone & EW \#2, AW \#1\\ 
    \hline
    Elizabeth    & Female & 82 & Caucasian & Alone& EW \#2\\ 
    \hline
    Lauren     & Female & 74 & African-American & in Community & EW \#3\\ 
    \hline
    Margaret   & Female & 82 & Caucasian & in Community & EW \#3\\ 
    \hline
    Cindy   & Female & 84 & Caucasian & in Community & EW \#3\\ 
    \hline
\end{tabular}}
\end{table}


\subsection{Phase 1: Exploratory Workshops}\label{process:ews}
We conducted three exploratory workshops to generate a broad spectrum of potential applications for on-body robots and to gather insights into the perceived benefits and barriers to adoption of these systems grounded in older adults' lived experiences. Each workshop lasted 70 to 90 minutes.


\textbf{Grounding.} Each session began with an introductory video about \textit{Calico}\footnote{link to video: \url{https://youtu.be/R1Mcj5uil6Q}}, followed by a demonstration of the design probe on a sleeve embedded with \textit{Calico}'s track system (see Fig. \ref{fig:teaser}). Interested participants wore the sleeve, and the robot was tele-operated along the track to allow them to experience the haptic feedback and on-body interaction firsthand. This hands-on introduction served as a foundation for the rest of the workshop. Then, we facilitated an open discussion to gather participants' initial impressions about on-body robots.

\textbf{Brainstorming.} Next, we facilitated an open-ended, collaborative map-making session to explore the design space for on-body robots aimed at enhancing the well-being of older adults. Participants first brainstormed potential use cases, writing their ideas on post-it notes and adding them to a shared mind map \cite{antony2023codesign} with images of common spaces for older adults such as park, bedroom \cite{chudyk2015destinations} to invoke imagination for usage. We then asked them to reflect on the potential benefits and challenges of adopting on-body robots, recording these insights on additional post-it notes and contributing to a second mind map. This process encouraged ongoing discussion, with participants organizing and visualizing emerging ideas collectively \cite{lee2017collaborative}.

\textbf{Preferences.} Finally, each participant selected the application ideas they found most promising, either for themselves or for their friends and family. For each selected idea, they shared key points explaining their choice, the factors that would influence their decision to use such a system, and any further thoughts on the robot's embodiment.

\subsection{Phase 1 Outcomes} \label{process:midpoint}
Phase 1 of our design process generated a diverse range of potential applications for on-body robots. Older adults envisioned use cases such as supporting activities of daily living (\eg personal hygiene, navigation), delivering targeted therapies like acupuncture, enhancing recreational activities (\eg dance), serving as wearable jewelry, mitigating fall risks, and enabling health monitoring and diagnostics. To explore this design space, we organized Phase 1 ideas along key dimensions—user movement, social context, and duration of use—highlighting differences and supporting systematic exploration. We also accounted for participants’ enthusiasm, for certain ideas (\eg walking). Building on this foundation, Phase 2 focused on three applications from Phase 1: Massage (recreation), Physical Therapy (rehabilitation), and Walking (daily living). These activities represented diverse aspects of older adults’ lives and varied in interaction styles, such as movement, duration, and social context, supporting a comprehensive exploration of the design space for on-body robots.



\subsection{Phase 2: Application-Focused Workshops} \label{process:dfws}

The application-focused workshops aimed to develop concrete designs for on-body robots tailored to a specific application area, providing a grounded perspective on interaction design of on-body robots. Each workshop took about 3 hours.

\textbf{Grounding.} We invited interested participants to try on the design probe showcasing locomotive and communicative capabilities. We, then, facilitated a collaborative map-making activity to explore the benefits of the application area, independent of the robot, to inform subsequent design activities.

\textbf{Contextualization.} Participants then imagined contexts for using on-body robots in the given applications, considering key factors like location, presence of others, and time of use. These ideas were then shared and discussed within the group.

\textbf{Interaction Flow.} Using free-form worksheets, participants brainstormed potential interaction designs for on-body robots in their chosen context, encouraging open-ended thinking. This was followed by three experience flow timelines\footnote{adapted from \href{https://tinyurl.com/jvrjr3nf}{experience-based co-design toolkit}}, focusing on the finer grained design---\textit{start, during,} and \textit{end}---of the interaction. A supplementary sheet outlining potential sensing and actuation capabilities, derived from the exploratory workshops, was provided to inspire the designs without limiting creativity. After completing each timeline, participants shared and discussed their envisioned designs with the group (see Supplementary Materials for worksheets).

\textbf{Bodystorming.} After a brief recess, participants engaged in a \textit{``bodystorming''} session, physically enacting their envisioned interactions with a 3D printed replica of the design probe. This activity helped anchor their designs in the practicalities of on-body interaction, offering valuable insights into the feasibility and user experience of their concepts \cite{segura2016bodystorming}.

\textbf{Personality and Embodiment.} Participants next designed the robot’s personality and embodiment using a set of worksheets \cite{axelsson2021social, cauchard2016emotion}. However, these worksheets were excluded from our analysis, as participants, fatigued at this late stage of the workshop, did not engage with them effectively.




\begin{figure*}[t]
\centering
\includegraphics[width=\textwidth]{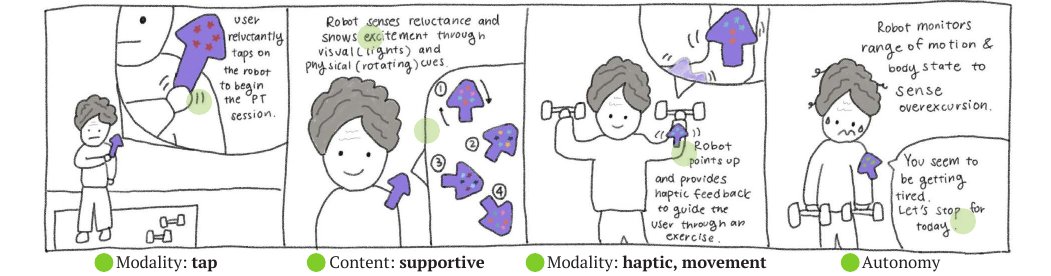}
\caption{Storyboard visualizing key components of the co-designed interaction of an on-body robot as a \textit{PT Coach}.}
\label{fig:storyboard-annotated}
\end{figure*}

\subsection{Data Analysis}

We transcribed workshop audio, digitized worksheets and mind maps, and extracted bodystorming session videos. Two researchers independently coded the data using a codebook\footnote{codebook and phase 1 data are provided in the \href{https://github.com/intuitivecomputing/Publications/blob/7f8b69042dbe6a1bd04598c66356058112780ecc/2025/HRI/Supplementary_2025_HRI_Antony_OnBody.pdf}{Supplementary Materials}} developed after an initial data review, resolving disagreements through discussion. Thematic analysis \cite{AMEEThematicAnalysis} revealed key ideas (\eg soft exterior, heart rate detection), which were grouped into higher-level concepts—design scopes and factors—through iterative team discussions. Finally, we mapped interactions between these concepts (units and anchors) to structure the initial design space for on-body robots.



\section{Co-Designed Interactions of On-Body Robots}

In the three application-focused workshops, participants designed on-body robots as \textit{Walking Sentinel}, \textit{Pet-Like Walking Companion}, \textit{Expert Masseuse} and \textit{Gamified PT Coach}. To highlight the diversity of interactions and roles envisioned, we present two co-designed applications from the Walking and Physical Therapy workshops.


\subsection{On-Body Robots as Fall Risk Mitigators}

Participants emphasized the critical impact of falls on older adults' quality of life. \textit{Melanie (F/76)} illustrated this sharing \pquotes{``Outside, inside, my head is down because I'm scared to death that I'm going to hit uneven pavement.''}. 





To address fall risks, participants conceptualized on-body robots as vigilant sentinels, providing active feedback to promote safe walking gaits. By delivering rhythmic cues—such as sound, vibration, or motion—similar to a gong or ocean waves, robots would prompt users to lift their feet while walking.

Participants also emphasized the importance of these robots detecting and responding to consistently poor gait patterns and environmental hazards (\eg curbs, overgrown roots). They envisioned the robots alerting users to imminent risks, with feedback that intensifies based on the proximity and severity of the danger, much like the escalating beeps of a car's backup sensor. Additionally, participants suggested that the robot could deploy countermeasures, such as activating lights in low-visibility conditions (\eg cloudy days, nighttime trips to the bathroom), further reducing the risk of falls.

This sentinel robot was envisioned to be worn and function continuously, seamlessly transitioning between active and passive modes based on the context (\eg user state, location).

Participants imagined wearing the robot near their ankles and, during bodystorming sessions, explored integrating it into footwear. The design was proposed to be utilitarian and discreet, ensuring minimal visibility---much like modern hearing aids---to mitigate any social stigma tied to its use.

\subsection{On-Body Robots As Physical Therapy Coach} 

Physical therapy (PT) was identified as a key component of healing holistically. To encourage older adults to consistently engage with PT protocols, participants envisioned on-body robots facilitating gamified physical therapy sessions. They emphasized making the experience enjoyable and feel as if \pquotes{``you are playing a game with your robot''} (see Fig. \ref{fig:storyboard-annotated})


To keep users engaged, participants imagined the on-body robot sensing the user’s state (\eg energy levels, mood) to provide optimal motivation; the robot would deliver dopamine-inducing, casino-style feedback throughout the session. This feedback was imagined in various modalities, including verbal (\eg \textit{``Good Job!''}), acoustic (\eg \textit{ding,ding,ding}), visual (\eg slot-machine like rainbow colors), and physical cues (\eg rotating on an axis). Additionally, the robot would track the user’s progress and incorporate it into its nudging behaviors to ensure adherence to the therapy regimen.

Participants cited overexertion and improper execution of exercises as major barriers to PT progress. As a solution, they envisioned the on-body robot monitoring the user's range of motion and exercise intensity, offering corrective feedback and adjusting the PT protocol as necessary. The robot was imagined to move across the body, providing feedback specific to the part being exercised at each stage. Initially, participants envisioned the robot providing visual feedback to guide exercises, but during bodystorming, they adapted this to a haptic-driven system after realizing that certain body poses prevented them from visually accessing the robot. 




\section{Characterizing the Design Space}

\begin{figure*}[t]
\centering
\includegraphics[width=\textwidth]{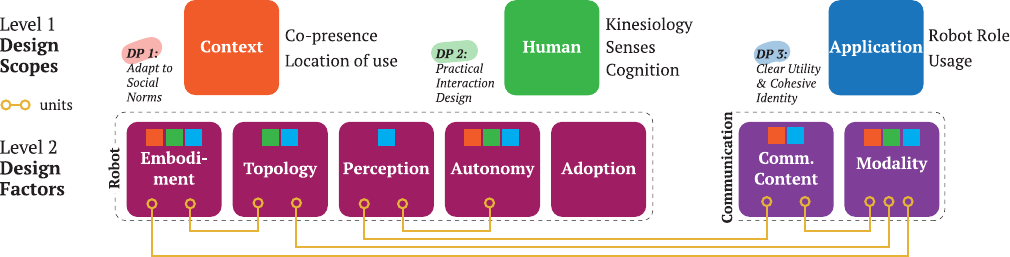}
\caption{We characterize an initial two-level design space for on-body robots consisting of design scopes and interconnected design factors.}
\label{fig:results-barriers-draft}
\end{figure*}


The possibilities envisioned in the exploratory workshops, combined with insights from the application-focused workshops, highlight the complexities of designing on-body robots for older adults. We characterize a two-level design space, composed of core concepts---design \textit{scopes} and design \textit{factors}---connected by design anchors and design units that synthesizes and organizes our workshop findings (see Fig. \ref{fig:results-barriers-draft}). The concepts on Level 1, namely \textit{context}, \textit{human}, and \textit{application}, scope the design space with the  design anchors establishing boundaries for the exploration of the key design factors of on-body robots around on the concepts in Level 2, \textit{robot} and \textit{communication}. The boundaries, defined by the design anchors, represent key  design principles derived from our workshops.  Design units illustrate the Level 2 design factors should be considered jointly as they interact closely.

\subsection{Design Scope: Context}
On-Body robots were envisioned to be used in diverse social contexts, defined by two key aspects: \textit{co-presence} (\ie presence of others) and \textit{location of use}.

\textbf{Co-presence.}
Participants imagined various actors being present during the use of on-body robots, including family members such as life partners (\eg spouses), children, and grandchildren; acquaintances such as sexual partners and friends; professionals like doctors, lawyers, physical therapists, occupational therapists, and caregivers; and even pets.
Participants also considered the impact of entities such as handbags, walking sticks, bikes on the interaction with the on-body robot.

\textbf{Location of use.} The envisioned usage locations for on-body robots ranged from fully public spaces like gyms, group classes, and beaches to semi-private areas such as hospital clinics and care facilities, and completely private spaces like bedrooms. The novelty of these robots can lead to heightened attention, especially in public contexts, where they may provoke strong reactions. For example, \textit{Cindy (F/84)} expressed how encountering an on-body robot in a public setting could challenge social norms, stating, \pquotes{``If we saw [\textbf{Calico}] today… in the city, [I would go] `what the hell?\dots what is that?'.''}


The context anchors downstream design decisions such as  maintaining privacy of communication content (\eg information about user's blood sugar level),  handling interactions between the robot  and the co-present entity (\eg cats startled by the robot) and  fitting form to social norms.


 \textbf{DP1:} \textit{Adapt to Social Norms.} On-Body robots, being novel and noticeable, may cause discomfort or embarrassment for older adults in social settings. To mitigate this, designs should be discreet, aesthetically pleasing, and context-aware. Subtle cues like gentle vibrations can maintain privacy in public, while private settings allow more expressive communication. Interaction with co-present social actors, such as the user's pets or caregivers, can help further normalize use. Social norms will evolve with wider adoption however early designs should align with current norms to facilitate acceptance.


\subsection{Design Scope: Human}

The unique proximity of on-body robots to the human body underscores the need to consider human factors in their design.

\textbf{Kinesiology.} A key consideration is the body's movement and posture in a usage context. Activity levels, ranging from sedentary to high-velocity movement, significantly influence the design space. The other aspect to consider is the posture of the human; in certain positions, the robot may not be reachable or be viewable to the user thus rendering certain modes of communication un-viable (\eg the \textit{Walking Sentinel} worn near the ankle, visual and verbal communication would fail).


Moreover, in certain postures, the user may unintentionally sit or lie on the robot, potentially causing discomfort unless the embodiment is designed with these scenarios in mind. \textit{Rachel (F/66)} raised a similar concern when thinking about the robot’s interaction in daily life, particularly in a sedentary setting: \pquotes{``I think of, of our friend [\textbf{Raymond}]… He spends a lot of time sitting in his recliner… How would this work in that kind of sitting environment? Do I just sit on the track?''}  Kinesiology anchors the robot’s embodiment, topology and communication modalities to the human form.

\textbf{Senses.} Designing communication pathways for on-body robots requires consideration of \textit{sensory factors}, such as variations in hearing, vision, and other sensory experiences.  Multiple communication modes are necessary to ensure accessibility and inclusivity, enabling effective interaction. \textit{Raymond (M/76)} highlighted this need, noting, \pquotes{``In terms of aging eyes, losing eye sight\dots There has to be more than one way of communicating with you ...''}.

\textbf{Cognition.} Certain \textit{cognitive factors}, \eg dementia or phobias like arachnophobia, may make on-body robots unsettling for some individuals. \textit{Rachel (F/66)} highlighted this concern, stating, \pquotes{``Something crawling on your body is a terrifying concept for somebody who’s genuinely aged now and may have less mobility \dots has some aspects of dementia where they’re not quite certain what’s happening around them.''}

 \textbf{DP2:} \textit{Practical Interaction Design.} Physical comfort is crucial for on-body robots due to their proximity to the body. These robots should be lightweight, soft, and adaptable to the user’s body, minimizing discomfort or injury. Designs must account for human kinesiology, avoiding interference with natural movement and being sensitive to areas prone to pain or discomfort. Robots should adjust behavior---such as pressure, movement, or positioning---based on physical cues. To prevent sensory overload, designers must consider the frequency and modality of feedback, ensuring it is gentle and non-intrusive.


\subsection{Design Scope: Application} 
The application the on-body robot facilitates for its user is central to its design; the application informs the role the robot assumes and the temporal aspects of its usage. 

\textbf{Role.} Participants envisioned various roles for on-body robots, including \textit{sentinel}, \textit{companion}, \textit{coach}, and \textit{tool}. Each of these roles influences the user’s level of reliance and trust, thereby defining the nature of the relationship and usage. For instance, \textit{Melanie (F/76)} highlighted how a deep reliance can form with the on-body robot, stating, \pquotes{``If it [on-body robot] is your lifeline, literally, then it’s kind of like [\textbf{Leona’s}] [SOS bracelet]. I would suspect you don’t ever turn that off.''}

\textbf{Usage.} Temporal interaction factors significantly shape the human-robot relationship. These include when and how often interactions occur, the duration of each interaction, and the long-term presence of the robot in user’s life. \textit{Melanie (F/76)} expressed the importance of how interactions begin and end: \pquotes{“If you do actually form a partnership or feel something about it, it would be nice if it wasn’t just… too abrupt, you know, like, hi, bye, not just turning and walking out the door.”} 


 The general interaction is shaped by the robot’s role and application, informing its design to establish clear utility and identity  \textbf{DP3:} \textit{Clear Utility and Cohesive Identity.} On-Body robots must justify their use by leveraging proximity and mobility to enable functions beyond those of stationary devices. As \textit{Lauren (F/74)} noted, ``\pquotes{I'd be waiting for not the calico [version] one o one… my watch does all of these things but it doesn't move… I don’t want it yet… while it looks like that.}'' A clear utility, coupled with a cohesive agency, will help build trust and create more natural, enjoyable interactions, ultimately making the robot a meaningful part of daily life.


\subsection{Design Factors: Robot}

\textbf{Embodiment.} The on-body nature of these robots makes their embodiment a crucial factor for user comfort and adoption, particularly for long-term use. Both the sensory aspects (\eg weight, softness) and visual elements (\eg colors, visibility) play a key role in shaping the user’s experience. Participants emphasized the embodiment should avoid signaling disability or medical conditions, which could discourage use. 

Participants emphasized the influence of the visual design and suggested enhancing “cuteness” while minimizing “weird” or “goofy” features to improve the robot’s appeal. Moreover, participants suggested aligning the robot's design with current fashion trends to promote usage, particularly in public spaces; \textit{Raymond (M/76)} remarked, \pquotes{“It might be more socially acceptable if it was something that is seen as an adornment instead of a monitor.”} Customizing the robot’s appearance, such as offering different colors or characters, could further encourage users to embrace the technology. As \textit{Randall (M/73)} noted, \pquotes{``You’d want an assortment of covers… you could put a ladybug or a little dog\dots whatever you want.''}

In terms of sensory factors, softness was repeatedly desired, as it could facilitate safer interactions for users with more sedentary lifestyles or in contexts like sleep aids. For instance, participants imagined the \textit{Expert Masseuse} robot having a soft exterior, allowing users to comfortably fall asleep during a session. Weight, on the other hand, was seen as a versatile feature. While lighter robots were preferred for designs like the \textit{Walking Sentinel} to avoid fatigue, heavier robots were considered beneficial for specific functions, such as adding resistance for exercise with the \textit{Gamified Coach} or applying additional pressure for massages with the \textit{Expert Masseuse}.  The embodiment has clear implications for robot topology. 


Participants explored the functional potential of robot embodiment. For instance, \textit{Randall (M/73)} envisioned a soft exterior serving as an airbag to reduce injury risks during falls or bumps. Beyond safety, a robot’s embodiment should align with its intended role and interaction style. The \textit{Walking Companion}, for example, was imagined to resemble a pet, reflecting the need for its design to match its relational and functional purpose. Embodiment can also enable communication modalities, as demonstrated by the arrow-shaped \textit{Gamified Coach} providing haptic and visual feedback (Fig. \ref{fig:storyboard-annotated}).
\textbf{Topology.} Participants envisioned a range of locations for the robot to be on body from their feet to their shoulders; from being worn on clothing such as a bolero or a sock to direct on-skin context. It is also important to consider the number of robots that need to be on a person at the same time for a given role. For instance, there might need to be one \textit{Walking Sentinel} for each leg, and multiple robots may be monitoring and providing feedback for different parts of the body for the \textit{Gamified Coach}. The location on body and the number of on-body robots inform the communication modality and must be considered to avoid any discomfort for the user.

\textbf{Perception.} On-Body robots were envisioned to monitor the user and the environment to inform their behaviors. In terms of environmental perception, participants imagined on-body robots perceiving factors such as location, potential hazards (\eg curbs, overhanging branches), weather conditions, visibility, temperature, and humidity. Accurate ego-location on the body was highlighted as a crucial feature, as it would guide both social and functional behaviors. For example, participants wanted the robot to adjust its behavior around sensitive areas (\eg neck), ensuring user comfort and safety.

Regarding the user, participants expected on-body robots to sense and interpret a range of human indicators such as emotions (\eg nervousness, anxiety), bio-statistics (\eg heart rate, oxygen levels, body temperature), physical states (\eg odor, muscle tension, movement), and goal progress.  The perception capabilities enable the necessary autonomy and the communication content necessary for the robot's role.

\textbf{Autonomy.} Leveraging their world and user model, on-body robots were expected to plan and execute actions towards their goals. Participants imagined the robot using a combination of explicit information-seeking and implicit sensing of real-time signals to adjust its plans to maintain trust and reliability.

Two primary classes of behaviors emerged during the workshops: reactive and proactive behaviors. Reactive behavior occurs in response to specific triggers, such as sensor recognition, or changes in the user’s body or environment. In contrast, proactive behaviors happen regularly as a method of checking in, requesting user input, or prompting the user as part of the robot’s functionality (\eg reminding the user to lift their foot or maintain correct posture). Tuning the frequency of behaviors based on the robot’s role, the severity of the event, and the user’s state and preferences were considered critical. 

Importantly, the degree of user input---whether frequent or occasional---would depend on the robot’s role and application. Moreover, participants envisioned shared autonomy paradigms where the robot's action would be influenced by inputs from external actors (\eg therapist) or would share information with caregivers (\eg medical professionals). However, participants emphasized that the user must always retain ultimate control. \textit{Norman (M/69)} underscored the importance of this dynamic, noting, \pquotes{``If it’s not good, [the user] asks the robot to stop and the robot stops immediately… then [the robot] looks at what’s going on and asks the person, ‘OK, I’m stopped. Do you want me to change the protocol or do you want me to just call it quits? You’re in control.’”} Robot responsiveness and user control was seen as vital for fostering usage and ensuring the robot’s actions remain aligned with the user’s preferences.

\textbf{Adoption.} Participants suggested including an onboarding process, featuring guided tutorials, to support initial adoption and lower the interaction learning curve. Participants also emphasized the need for convenience in daily tasks like charging, cleaning, and storing the robot. Designing the robot with easy-to-clean or washable materials is essential for maintaining hygiene, especially for a device in constant contact with the body. Participants noted that older adults often misplace small devices, so features like a ``find my robot'' function or the robot autonomously returning to its docking station were suggested to help mitigate this issue. Additionally, extended battery life were desired to enhance convenience by enabling more widespread usage. All \textit{design scopes} influence but do not impose clear constraints on the design for adoption.

\subsection{Design Factors: Communication}

\textbf{Modality.} Participants envisioned communication with on-body robots using verbal and non-verbal methods. They favored natural, conversational verbal interactions over commands. Non-verbal modes included gestures (e.g., two fingers touching), physical actions (e.g., tapping, pressing or slapping the robot), and remote controls for specific scenarios.

Participants imagined a wide range of robot-to-user communication methods, taking advantage of the robot’s proximity to the body and its mobility. \textit{Physical movement} in meaningful patterns or navigating to specific body areas was proposed to convey intent, affect, or information. \textit{Visual feedback}, such as blinking or color-changing LEDs, was another suggested option; however, designing on-robots to convey more complex information via visual feedback was thought to be challenging.

\textit{Haptic feedback}, both kinesthetic (\eg tugs, pinches) and vibro-tactile (\eg buzzes), was widely discussed. \textit{Raymond (M/76)} noted its importance for visually impaired users, explaining, \pquotes{“It could literally be in the shape of an arrow… you could feel it if it actually moved…”} Participants also raised the idea of \textit{olfactory output}, where the robot could emit essential oils or burn incense to provide feedback through scent, opening up novel sensory communication avenues.

For \textit{verbal feedback}, thoughtful voice design characterized as ‘cute,’ and ‘lovely’ was emphasized; moreover, communicating via hearing aids was suggested as a way to improve communication for users with hearing impairments or in noisy environments. \textit{Non-verbal acoustic cues}, like “ding ding ding,” were also imagined for certain contexts.

The close proximity of the robot to the body opens up new opportunities for communication beyond traditional verbal interaction. To ensure clear and intuitive communication, participants stressed the importance of building on familiar conventions, such as the colors of traffic lights, to create easily understandable patterns of feedback underscoring  the need for the modality to support the content of communication.

Participants designed multi-modal affect communication for on-body robots, particularly in social roles. For example, the \textit{Walking Companion} was imagined to light up and use acoustic cues to express excitement before a walk, while the \textit{Gamified Coach} would signal the end of a session by performing a playful dance, including spinning in place. Leveraging multi-sensory experiences could help establish the robot’s character and foster deeper emotional connections with users, ultimately reinforcing adoption and continued usage \cite{duffy2003anthropomorphism}. However, participants noted the importance of balancing these communication pathways to prevent sensory overload to ensure the robot integrates smoothly into the user’s daily life.

\textbf{Communication Content.}
Two types of content for communication emerged: supportive and informative. Supportive content provides encouragement or rewards, offering “dopamine” feedback to motivate the user or celebrate successful events. Informative content, on the other hand, includes data from the robot’s sensors—such as alerts to potential dangers—or information crucial to the user, like reminders, explanations of robot actions, or corrections (\eg adjusting the user’s posture or addressing discomfort causing robot action).

\section{Using the Design Space}

Our design space serves as an initial framework for research and practical development in the field of on-body robots, particularly for older adults. It can be leveraged in two distinct ways: first, to identify and investigate open research questions, and second, to guide the creation of on-body robot prototypes. Our design space aims to enable an iterative exploration of on-body robots as a practical HRI paradigm for older adults.

\subsection{Open-Questions for On-Body Robots}

The design space helps uncover several open research questions critical to advancing on-body robots as a technology. At the core of these questions are the design factors articulated in Level 2 of the design space: robot (embodiment, topology, perception, autonomy, and adoption) and communication (content and modality). For instance, \textit{How can on-body robots use olfactory output to communicate? How to account for the fragility of older adults' skins? What are the trade-offs between using a single on-body robot and multiple robots distributed across the body?
} These questions highlight the need for deeper exploration into how each concept can be practically implemented thereby refining this initial design space.

Understanding the interplay within design units presents another layer of inquiry. For example, \textit{How to design embodiments suitable for multiple on-body robots without overwhelming users?} There are also broader questions on how to practically achieve the presented design principles for different populations and settings. These open questions can guide future work to enhance on-body robots’ real-world viability and refine the design space through situated co-design \cite{stegner2023situated}.

\subsection{Building On-Body Robot Prototypes}

Our design space offers a structured approach for creating functional prototypes of on-body robots facilitating further exploration of this HRI paradigm. The process begins by defining the design scopes at Level 1, which covers the target population (\eg older adults, blind or visually impaired (BVI) individuals), the robot’s application (\eg navigation, acupuncture), and the expected context of use (\eg clinics, park). These foundational decisions establish clear boundaries and constraints for the robot’s design, ensuring that it aligns with user needs and environmental factors. For instance, while designing for BVI people, visual modalities cannot be used.

Next, the design factors---such as the robot’s embodiment, communication modalities, and autonomy---can be explored broadly within these established boundaries. For instance, acoustics and robot movement can be evaluated as communication modality for BVI individuals. The nascent nature of on-body robots requires significant exploration with functional prototypes, and iterative testing with users to garner better understanding of this design space. For example, different embodiments can be built and evaluated for their fit with social norms, while communication pathways may be adapted to include affordances tailored to certain populations.

Using our design space, researchers and designers can find a set of feasible prototypes for exploring relationships between design elements and broader open questions in this novel HRI space (see Supplementary Materials for our visual guide).

\section{Reflections on Co-Designing On-Body Robots}
The sensitive and safety-critical applications envisioned for on-body robots (\eg fall prevention) highlight the importance of involving end-users as design partners to ensure usability and adoption. Our design workshops provided key insights for effectively engaging older adults as co-designers.

\subsubsection{Lived Experiences with On-Body Robots} The novelty of on-body robots underscores the need to introduce these concepts in an experiential and digestible manner to engage participants as effective design partners \cite{ostrowski2021personal, ostrowski2021long,lee2017steps,lee2018reframing}. Hands-on demonstration of our design probe jump-started participants’ design thinking and helped demystify on-body robots. We also observed that participants who engaged in both exploratory and application-focused workshops were more comfortable imagining interaction paradigms compared to those who only participated in the later. Introducing on-body robots with a design probe, with time to reflect between sessions, encouraged more active and creative engagement \cite{mahmood2024our}. Thus, a multi-stage design process, with shorter, focused workshops may enable deeper involvement from older adults.

\subsubsection{Structure in Design} 
Utilizing the experience flow worksheets introduced a malleable structure into the design activity, making it less overwhelming for older adults to engage with the design process. Moreover, conducting a free-form activity, followed by more structured experience flows, helped engage participants’ creativity while simultaneously making the design process more approachable—particularly given the novelty and complexity of on-body robots.

\subsubsection{Bodystorming On-Body Interactions} Bodystorming played a pivotal role in our design process by allowing participants to identify subtle, grounded design considerations \cite{oulasvirta2003understanding, schleicher2010bodystorming, stegner2023situated} and engage more deeply with the physical aspects of the interaction \cite{segura2016bodystorming}. For example, \textit{Raymond} bodystormed the \textit{Physical Therapy} scenario from the perspective of a blind person, highlighting the importance of multimodal communication. For the \textit{Massage} scenario, bodystorming prompted participants to consider how the robot would adapt to different body areas, sparking discussions on custom 3D-printed form factors tailored to individual users and specific therapies.


\section{Limitations and Future Work}
Our design process uncovered promising applications and provided valuable insights into the design space of on-body robots. However, to gain a deeper, more grounded understanding of this interaction paradigm, future work should implement the proposed interactions for on-body robots and engage older adults in evaluation processes to further refine the design space. Additional workshops using on-body robots beyond Calico \cite{sathya2022calico} could provide further insights. The co-design partners involved in this study are not fully representative of the diverse population of older adults, who vary widely in physical and cognitive abilities. Future research should involve a broader spectrum of older adults in the design process to explore the appropriateness and specific design needs of on-body robots for different subgroups within the aging population.


\section*{Acknowledgment}
This work was supported in part by the JHU Malone Center for Engineering in Healthcare.



\clearpage
\balance

\bibliographystyle{IEEEtran}
\bibliography{references}

\begin{thebibliography}{10}
\providecommand{\url}[1]{#1}
\csname url@samestyle\endcsname
\providecommand{\newblock}{\relax}
\providecommand{\bibinfo}[2]{#2}
\providecommand{\BIBentrySTDinterwordspacing}{\spaceskip=0pt\relax}
\providecommand{\BIBentryALTinterwordstretchfactor}{4}
\providecommand{\BIBentryALTinterwordspacing}{\spaceskip=\fontdimen2\font plus
\BIBentryALTinterwordstretchfactor\fontdimen3\font minus \fontdimen4\font\relax}
\providecommand{\BIBforeignlanguage}[2]{{%
\expandafter\ifx\csname l@#1\endcsname\relax
\typeout{** WARNING: IEEEtran.bst: No hyphenation pattern has been}%
\typeout{** loaded for the language `#1'. Using the pattern for}%
\typeout{** the default language instead.}%
\else
\language=\csname l@#1\endcsname
\fi
#2}}
\providecommand{\BIBdecl}{\relax}
\BIBdecl

\bibitem{mulrow1990quality}
C.~D. Mulrow, C.~Aguilar, J.~E. Endicott, M.~R. Tuley, R.~Velez, W.~S. Charlip, M.~C. Rhodes, J.~A. Hill, and L.~A. DeNino, ``Quality-of-life changes and hearing impairment: a randomized trial,'' \emph{Annals of internal medicine}, vol. 113, no.~3, pp. 188--194, 1990.

\bibitem{litchman2017real}
M.~L. Litchman and N.~A. Allen, ``Real-time continuous glucose monitoring facilitates feelings of safety in older adults with type 1 diabetes: a qualitative study,'' \emph{Journal of diabetes science and technology}, vol.~11, no.~5, pp. 988--995, 2017.

\bibitem{chung2023community}
J.~Chung, H.~R. Brakey, B.~Reeder, O.~Myers, and G.~Demiris, ``Community-dwelling older adults' acceptance of smartwatches for health and location tracking,'' \emph{International journal of older people nursing}, vol.~18, no.~1, p. e12490, 2023.

\bibitem{kim2011epidermalelectronics}
\BIBentryALTinterwordspacing
D.-H. Kim, N.~Lu, R.~Ma, Y.-S. Kim, R.-H. Kim, S.~Wang, J.~Wu, S.~M. Won, H.~Tao, A.~Islam, K.~J. Yu, T.~il~Kim, R.~Chowdhury, M.~Ying, L.~Xu, M.~Li, H.-J. Chung, H.~Keum, M.~McCormick, P.~Liu, Y.-W. Zhang, F.~G. Omenetto, Y.~Huang, T.~Coleman, and J.~A. Rogers, ``Epidermal electronics,'' \emph{Science}, vol. 333, no. 6044, pp. 838--843, 2011. [Online]. Available: \url{https://www.science.org/doi/abs/10.1126/science.1206157}
\BIBentrySTDinterwordspacing

\bibitem{weigel2015iskin}
M.~Weigel, T.~Lu, G.~Bailly, A.~Oulasvirta, C.~Majidi, and J.~Steimle, ``Iskin: flexible, stretchable and visually customizable on-body touch sensors for mobile computing,'' in \emph{Proceedings of the 33rd Annual ACM Conference on Human Factors in Computing Systems}, 2015, pp. 2991--3000.

\bibitem{kao2015nailo}
H.-L. Kao, A.~Dementyev, J.~A. Paradiso, and C.~Schmandt, ``Nailo: fingernails as an input surface,'' in \emph{Proceedings of the 33rd Annual ACM Conference on Human Factors in Computing Systems}, 2015, pp. 3015--3018.

\bibitem{holz2012implanted}
C.~Holz, T.~Grossman, G.~Fitzmaurice, and A.~Agur, ``Implanted user interfaces,'' in \emph{Proceedings of the SIGCHI conference on human factors in computing systems}, 2012, pp. 503--512.

\bibitem{poupyrev2016project}
I.~Poupyrev, N.-W. Gong, S.~Fukuhara, M.~E. Karagozler, C.~Schwesig, and K.~E. Robinson, ``Project jacquard: interactive digital textiles at scale,'' in \emph{Proceedings of the 2016 CHI Conference on Human Factors in Computing Systems}, 2016, pp. 4216--4227.

\bibitem{vega2014beauty}
K.~Vega and H.~Fuks, ``Beauty technology: body surface computing,'' \emph{Computer}, vol.~47, no.~4, pp. 71--75, 2014.

\bibitem{orlov2012technology}
L.~M. Orlov, ``Technology for aging in place,'' \emph{Aging in Place Technology Watch}, 2012.

\bibitem{fournier2020designing}
H.~Fournier, I.~Kondratova, and H.~Molyneaux, ``Designing digital technologies and safeguards for improving activities and well-being for aging in place,'' in \emph{HCI International 2020--Late Breaking Papers: Universal Access and Inclusive Design: 22nd HCI International Conference, HCII 2020, Copenhagen, Denmark, July 19--24, 2020, Proceedings 22}.\hskip 1em plus 0.5em minus 0.4em\relax Springer, 2020, pp. 524--537.

\bibitem{kong2006design}
K.~Kong and D.~Jeon, ``Design and control of an exoskeleton for the elderly and patients,'' \emph{IEEE/ASME Transactions on mechatronics}, vol.~11, no.~4, pp. 428--432, 2006.

\bibitem{jayaraman2022modular}
C.~Jayaraman, K.~R. Embry, C.~K. Mummidisetty, Y.~Moon, M.~Giffhorn, S.~Prokup, B.~Lim, J.~Lee, Y.~Lee, M.~Lee \emph{et~al.}, ``Modular hip exoskeleton improves walking function and reduces sedentary time in community-dwelling older adults,'' \emph{Journal of neuroengineering and rehabilitation}, vol.~19, no.~1, p. 144, 2022.

\bibitem{jung2019older}
M.~M. Jung and G.~D. Ludden, ``What do older adults and clinicians think about traditional mobility aids and exoskeleton technology?'' \emph{ACM Transactions on Human-Robot Interaction (THRI)}, vol.~8, no.~2, pp. 1--17, 2019.

\bibitem{sathya2022calico}
\BIBentryALTinterwordspacing
A.~Sathya, J.~Li, T.~Rahman, G.~Gao, and H.~Peng, ``Calico: Relocatable on-cloth wearables with fast, reliable, and precise locomotion,'' \emph{Proc. ACM Interact. Mob. Wearable Ubiquitous Technol.}, vol.~6, no.~3, Sep. 2022. [Online]. Available: \url{https://doi.org/10.1145/3550323}
\BIBentrySTDinterwordspacing

\bibitem{dementyev2017skinbot}
A.~Dementyev, J.~Hernandez, S.~Follmer, I.~Choi, and J.~Paradiso, ``Skinbot: A wearable skin climbing robot,'' in \emph{Adjunct Proceedings of the 30th Annual ACM Symposium on User Interface Software and Technology}, 2017, pp. 5--6.

\bibitem{dementyev2016rovables}
A.~Dementyev, H.-L. Kao, I.~Choi, D.~Ajilo, M.~Xu, J.~A. Paradiso, C.~Schmandt, and S.~Follmer, ``Rovables: Miniature on-body robots as mobile wearables,'' in \emph{Proceedings of the 29th Annual Symposium on User Interface Software and Technology}, 2016, pp. 111--120.

\bibitem{hall1990hidden}
\BIBentryALTinterwordspacing
E.~Hall, \emph{The Hidden Dimension}.\hskip 1em plus 0.5em minus 0.4em\relax Knopf Doubleday Publishing Group, 1990. [Online]. Available: \url{https://books.google.com/books?id=zGYPwLj2dCoC}
\BIBentrySTDinterwordspacing

\bibitem{muller1993participatory}
M.~J. Muller and S.~Kuhn, ``Participatory design,'' \emph{Communications of the ACM}, vol.~36, no.~6, pp. 24--28, 1993.

\bibitem{rogers2022maximizing}
W.~A. Rogers, T.~Kadylak, and M.~A. Bayles, ``Maximizing the benefits of participatory design for human--robot interaction research with older adults,'' \emph{Human factors}, vol.~64, no.~3, pp. 441--450, 2022.

\bibitem{antony2023codesign}
V.~N. Antony, S.~M. Cho, and C.-M. Huang, ``Co-designing with older adults, for older adults: robots to promote physical activity,'' in \emph{Proceedings of the 2023 ACM/IEEE International Conference on Human-Robot Interaction}, 2023, pp. 506--515.

\bibitem{oneill2017exoshoulder}
C.~T. O'Neill, N.~S. Phipps, L.~Cappello, S.~Paganoni, and C.~J. Walsh, ``A soft wearable robot for the shoulder: Design, characterization, and preliminary testing,'' in \emph{2017 International Conference on Rehabilitation Robotics (ICORR)}.\hskip 1em plus 0.5em minus 0.4em\relax IEEE, 2017, pp. 1672--1678.

\bibitem{gao2023portable}
W.~Gao, A.~Di~Lallo, and H.~Su, ``A portable powered soft exoskeleton for shoulder assistance during functional movements: Design and evaluation,'' in \emph{International Symposium on Medical Robotics (ISMR)}, 2023.

\bibitem{in2015exo}
H.~In, B.~B. Kang, M.~Sin, and K.-J. Cho, ``Exo-glove: A wearable robot for the hand with a soft tendon routing system,'' \emph{IEEE Robotics \& Automation Magazine}, vol.~22, no.~1, pp. 97--105, 2015.

\bibitem{huo2014exolower}
W.~Huo, S.~Mohammed, J.~C. Moreno, and Y.~Amirat, ``Lower limb wearable robots for assistance and rehabilitation: A state of the art,'' \emph{IEEE systems Journal}, vol.~10, no.~3, pp. 1068--1081, 2014.

\bibitem{asbeck2014stronger}
A.~T. Asbeck, S.~M. De~Rossi, I.~Galiana, Y.~Ding, and C.~J. Walsh, ``Stronger, smarter, softer: next-generation wearable robots,'' \emph{IEEE Robotics \& Automation Magazine}, vol.~21, no.~4, pp. 22--33, 2014.

\bibitem{zhou2019novel}
H.~Zhou, A.~Mohammadi, D.~Oetomo, and G.~Alici, ``A novel monolithic soft robotic thumb for an anthropomorphic prosthetic hand,'' \emph{IEEE Robotics and Automation Letters}, vol.~4, no.~2, pp. 602--609, 2019.

\bibitem{shafti2021playing}
A.~Shafti, S.~Haar, R.~Mio, P.~Guilleminot, and A.~A. Faisal, ``Playing the piano with a robotic third thumb: assessing constraints of human augmentation,'' \emph{Scientific reports}, vol.~11, no.~1, p. 21375, 2021.

\bibitem{muehlhaus2023need}
M.~Muehlhaus, M.~Koelle, A.~Saberpour, and J.~Steimle, ``I need a third arm! eliciting body-based interactions with a wearable robotic arm,'' in \emph{Proceedings of the 2023 CHI Conference on Human Factors in Computing Systems}, 2023, pp. 1--15.

\bibitem{vatsal2017wearing}
V.~Vatsal and G.~Hoffman, ``Wearing your arm on your sleeve: Studying usage contexts for a wearable robotic forearm,'' in \emph{2017 26th IEEE International Symposium on Robot and Human Interactive Communication (RO-MAN)}.\hskip 1em plus 0.5em minus 0.4em\relax IEEE, 2017, pp. 974--980.

\bibitem{tsumaki201220}
Y.~Tsumaki, F.~Ono, and T.~Tsukuda, ``The 20-dof miniature humanoid mh-2: A wearable communication system,'' in \emph{2012 IEEE International Conference on Robotics and Automation}.\hskip 1em plus 0.5em minus 0.4em\relax IEEE, 2012, pp. 3930--3935.

\bibitem{jiang2019survey}
H.~Jiang, S.~Lin, V.~Prabakaran, M.~R. Elara, and L.~Sun, ``A survey of users' expectations towards on-body companion robots,'' in \emph{Proceedings of the 2019 on Designing Interactive Systems Conference}, 2019, pp. 621--632.

\bibitem{birkmeyer2011clash}
P.~Birkmeyer, A.~G. Gillies, and R.~S. Fearing, ``Clash: Climbing vertical loose cloth,'' in \emph{2011 IEEE/RSJ International Conference on Intelligent Robots and Systems}.\hskip 1em plus 0.5em minus 0.4em\relax IEEE, 2011, pp. 5087--5093.

\bibitem{alves2015social}
P.~Alves-Oliveira, S.~Petisca, F.~Correia, N.~Maia, and A.~Paiva, ``Social robots for older adults: Framework of activities for aging in place with robots,'' in \emph{Social Robotics: 7th International Conference, ICSR 2015, Paris, France, October 26-30, 2015, Proceedings 7}.\hskip 1em plus 0.5em minus 0.4em\relax Springer, 2015, pp. 11--20.

\bibitem{ostrowski2021personal}
A.~K. Ostrowski, C.~N. Harrington, C.~Breazeal, and H.~W. Park, ``Personal narratives in technology design: the value of sharing older adults’ stories in the design of social robots,'' \emph{Frontiers in Robotics and AI}, vol.~8, p. 716581, 2021.

\bibitem{alves2021children}
P.~Alves-Oliveira, P.~Arriaga, A.~Paiva, and G.~Hoffman, ``Children as robot designers,'' in \emph{Proceedings of the 2021 ACM/IEEE International Conference on Human-Robot Interaction}, 2021, pp. 399--408.

\bibitem{nanavati2023design}
A.~Nanavati, P.~Alves-Oliveira, T.~Schrenk, E.~K. Gordon, M.~Cakmak, and S.~S. Srinivasa, ``Design principles for robot-assisted feeding in social contexts,'' in \emph{Proceedings of the 2023 ACM/IEEE International Conference on Human-Robot Interaction}, 2023, pp. 24--33.

\bibitem{lee2018reframing}
H.~R. Lee and L.~D. Riek, ``Reframing assistive robots to promote successful aging,'' \emph{ACM Transactions on Human-Robot Interaction (THRI)}, vol.~7, no.~1, pp. 1--23, 2018.

\bibitem{moharana2019robots}
S.~Moharana, A.~E. Panduro, H.~R. Lee, and L.~D. Riek, ``Robots for joy, robots for sorrow: community based robot design for dementia caregivers,'' in \emph{2019 14th ACM/IEEE International Conference on Human-Robot Interaction (HRI)}.\hskip 1em plus 0.5em minus 0.4em\relax IEEE, 2019, pp. 458--467.

\bibitem{lee2017steps}
H.~R. Lee, S.~{\v{S}}abanovi{\'c}, W.-L. Chang, S.~Nagata, J.~Piatt, C.~Bennett, and D.~Hakken, ``Steps toward participatory design of social robots: mutual learning with older adults with depression,'' in \emph{Proceedings of the 2017 ACM/IEEE international conference on human-robot interaction}, 2017, pp. 244--253.

\bibitem{harrington2018designing}
C.~N. Harrington, L.~Wilcox, K.~Connelly, W.~Rogers, and J.~Sanford, ``Designing health and fitness apps with older adults: Examining the value of experience-based co-design,'' in \emph{Proceedings of the 12th EAI international conference on pervasive computing technologies for healthcare}, 2018, pp. 15--24.

\bibitem{nevay2015role}
S.~Nevay and C.~S. Lim, ``The role of co-design in wearables adoption,'' in \emph{Contemporary Ergonomics and Human Factors 2015: Proceedings of the International Conference on Ergonomics \& Human Factors}, 2015.

\bibitem{bjorling2019participatory}
E.~A. Bj{\"o}rling and E.~Rose, ``Participatory research principles in human-centered design: engaging teens in the co-design of a social robot,'' \emph{Multimodal Technologies and Interaction}, vol.~3, no.~1, p.~8, 2019.

\bibitem{axelsson2021social}
M.~Axelsson, R.~Oliveira, M.~Racca, and V.~Kyrki, ``Social robot co-design canvases: A participatory design framework,'' \emph{ACM Transactions on Human-Robot Interaction (THRI)}, vol.~11, no.~1, pp. 1--39, 2021.

\bibitem{ostrowski2021long}
A.~K. Ostrowski, C.~Breazeal, and H.~W. Park, ``Long-term co-design guidelines: empowering older adults as co-designers of social robots,'' in \emph{2021 30th IEEE International Conference on Robot \& Human Interactive Communication (RO-MAN)}.\hskip 1em plus 0.5em minus 0.4em\relax IEEE, 2021, pp. 1165--1172.

\bibitem{hutchinson2003technology}
H.~Hutchinson, W.~Mackay, B.~Westerlund, B.~B. Bederson, A.~Druin, C.~Plaisant, M.~Beaudouin-Lafon, S.~Conversy, H.~Evans, H.~Hansen \emph{et~al.}, ``Technology probes: inspiring design for and with families,'' in \emph{Proceedings of the SIGCHI conference on Human factors in computing systems}, 2003, pp. 17--24.

\bibitem{gough2021co}
P.~Gough, A.~B. Kocaballi, K.~Z. Naqshbandi, K.~Cochrane, K.~Mah, A.~G. Pillai, Y.~Yorulmaz, A.~K. Deny, and N.~Ahmadpour, ``Co-designing a technology probe with experienced designers,'' in \emph{Proceedings of the 33rd Australian Conference on Human-Computer Interaction}, 2021, pp. 1--13.

\bibitem{chudyk2015destinations}
A.~M. Chudyk, M.~Winters, M.~Moniruzzaman, M.~C. Ashe, J.~S. Gould, and H.~McKay, ``Destinations matter: The association between where older adults live and their travel behavior,'' \emph{Journal of Transport \& Health}, vol.~2, no.~1, pp. 50--57, 2015.

\bibitem{lee2017collaborative}
H.~R. Lee, S.~{\v{S}}abanovi{\'c}, and S.~S. Kwak, ``Collaborative map making: A reflexive method for understanding matters of concern in design research,'' in \emph{Proceedings of the 2017 CHI Conference on Human Factors in Computing Systems}, 2017, pp. 5678--5689.

\bibitem{segura2016bodystorming}
E.~Segura, L.~Vidal, and A.~Rostami, ``Bodystorming for movement-based interaction design,'' \emph{Human Technology}, vol.~12, no.~2, pp. 193--251, 2016.

\bibitem{cauchard2016emotion}
J.~R. Cauchard, K.~Y. Zhai, M.~Spadafora, and J.~A. Landay, ``Emotion encoding in human-drone interaction,'' in \emph{2016 11th ACM/IEEE International Conference on Human-Robot Interaction (HRI)}.\hskip 1em plus 0.5em minus 0.4em\relax IEEE, 2016, pp. 263--270.

\bibitem{AMEEThematicAnalysis}
M.~Kiger and L.~Varpio, ``Thematic analysis of qualitative data: Amee guide no. 131,'' \emph{Medical Teacher}, vol.~42, pp. 1--9, 05 2020.

\bibitem{duffy2003anthropomorphism}
B.~R. Duffy, ``Anthropomorphism and the social robot,'' \emph{Robotics and autonomous systems}, vol.~42, no. 3-4, pp. 177--190, 2003.

\bibitem{stegner2023situated}
L.~Stegner, E.~Senft, and B.~Mutlu, ``Situated participatory design: A method for in situ design of robotic interaction with older adults,'' in \emph{Proceedings of the 2023 CHI Conference on Human Factors in Computing Systems}, 2023, pp. 1--15.

\bibitem{mahmood2024our}
A.~Mahmood and C.-M. Huang, ``From our lab to their homes: Learnings from longitudinal field research with older adults,'' \emph{arXiv preprint arXiv:2409.15495}, 2024.

\bibitem{oulasvirta2003understanding}
A.~Oulasvirta, E.~Kurvinen, and T.~Kankainen, ``Understanding contexts by being there: case studies in bodystorming,'' \emph{Personal and ubiquitous computing}, vol.~7, pp. 125--134, 2003.

\bibitem{schleicher2010bodystorming}
D.~Schleicher, P.~Jones, and O.~Kachur, ``Bodystorming as embodied designing,'' \emph{interactions}, vol.~17, no.~6, pp. 47--51, 2010.

\end{thebibliography}

\end{document}


\title{The Design of On-Body Robots for Older Adults: 
Supplementary Materials
}

\author{Victor Nikhil Antony$^{1}$, Clara Jeon$^{1}$, Jiasheng Li$^{2}$, Ge Gao$^{2}$, \\ Huaishu Peng$^{2}$, Anastasia K. Ostrowski$^{3}$, and Chien-Ming Huang$^{1}$
\thanks{$^{1}$Johns Hopkins University, Baltimore, Maryland, USA {\tt\small \{vantony1, cjeon6, chienming.huang\}@jhu.edu}. $^{2}$University of Maryland, College Park, Maryland, USA {\tt\small \{jsli, gegao, huaishu\}@umd.edu}. $^{3}$Purdue University, West Lafayette, IN, USA {\tt\small akostrow@purdue.edu}.}}

\maketitle

\begin{figure*}
\centering
\includegraphics[width=\textwidth]{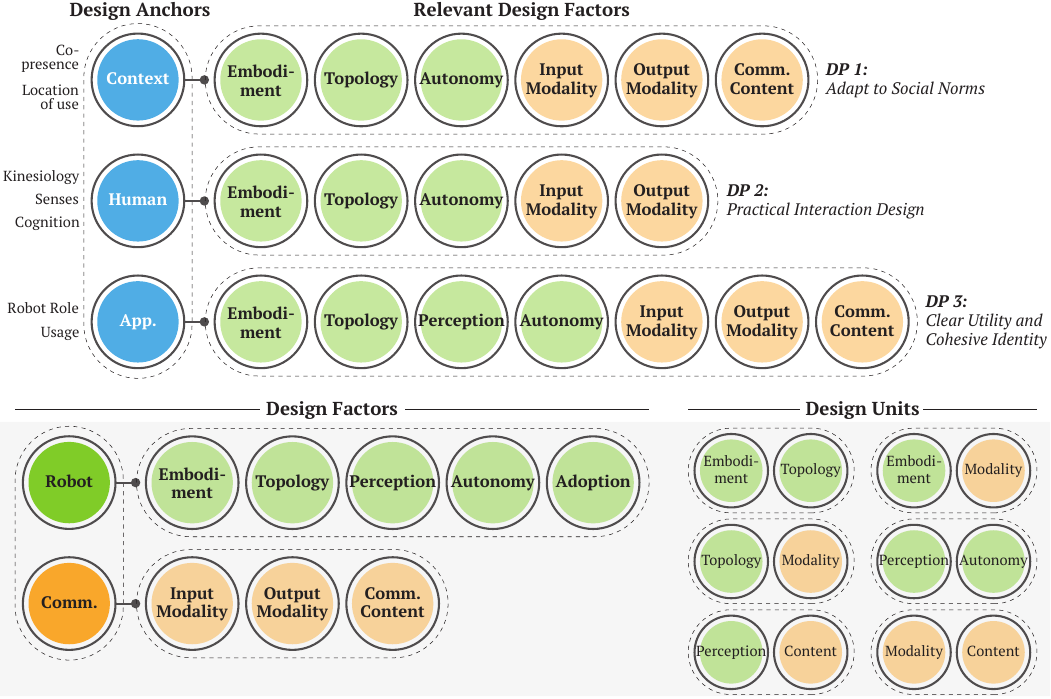}
\caption{A visual guide to enable researchers and designers working on on-body robots to easily use our design space.}
\label{fig:results-barriers-draft}
\end{figure*}

\begin{table*}[ht!]
\centering
\caption{Organization of Phase 1 outputs along movement dimensions}
\begin{tabular}{|m{0.235\linewidth}|m{0.229\linewidth}|m{0.229\linewidth}|m{0.246\linewidth}|} 
\hline
 & \textbf{User High Movement} & \textbf{User Medium Movement} & \textbf{User Low Movement} \\ 
\hline
\textbf{Robot High Movement} & 
Home Exercise & 
 & 
Picking up Items \\ 
\hline
\textbf{Robot Medium Movement} & 
Posture Correction\par{}Walking\par{}Gait Correction\par{}Home Exercise & 
Posture Correction\par{}Physical Therapy\par{}Walking\par{}Home Exercise & 
Posture Correction\par{}Yoga\par{}Home Exercise\par{}Acupressure/Acupuncture\par{}Massage\par{}Topical Medicine Application\par{}Use in Clinics and Hospitals\par{}Gardening \\ 
\hline
\textbf{Robot Low Movement} & 
Fall Risk Mitigation\par{}Collision Alert\par{}Walking\par{}Gait Correction\par{}Home Exercise\par{}Fashion Use (Jewelry)\par{}Reminders\par{}Balance Tracking\par{}Track Person\par{}Health Diagnostics/Sensing & 
Fall Risk Mitigation\par{}Physical Therapy\par{}Home Exercise\par{}Walking\par{}Fashion Use (Jewelry)\par{}Reminders\par{}Balance Tracking\par{}Track Person\par{}Health Diagnostics/Sensing & 
Fall Risk Mitigation\par{}Yoga\par{}Home Exercise\par{}Fashion Use (Jewelry)\par{}Meditation\par{}Reminders\par{}Acupressure/Acupuncture\par{}Massage\par{}Topical Medicine Application\par{}Balance Tracking\par{}Track Person\par{}Chew Assistance\par{}Gardening\par{}Water Consumption Tracker\par{}Feedback on Self Odor \\ 
\hline
\end{tabular}
\end{table*}

\begin{table*}[ht!]
\centering
\caption{Organization of Phase 1 applications along Robot Time on User and Co-presense Context Dimensions}
\begin{tabular}{|p{0.3\linewidth}|p{0.7\linewidth}|} 
\hline
\textbf{Category} & \textbf{Details} \\ 
\hline
\textbf{Robot Time on User} & \\ 
\hline
\textit{During Application} & 
Home Exercise, Physical Therapy, Yoga, Acupressure/Acupuncture, Massage, Topical Medicine Application, Gardening, Meditation, Use in Clinics and Hospitals, Chew Assistance \\ 
\hline
\textit{All the Time} & 
Posture Correction, Fall Risk Mitigation, Collision Alert, Gait Correction, Walking, Balance Tracking, Track Person, Health Diagnostics/Sensing, Picking up Items, Reminders, Fashion Use (Jewelry), Water Consumption Tracker, Feedback on Self Odor \\ 
\hline
\textbf{Social Interaction Context} & \\ 
\hline
\textit{Individual} & 
Home Exercise, Yoga, Physical Therapy, Acupressure/Acupuncture, Massage, Topical Medicine Application, Gardening, Meditation, Walking, Chew Assistance, Gait Correction, Health Diagnostics/Sensing \\ 
\hline
\textit{Experts} & 
Use in Clinics and Hospitals, Physical Therapy, Meditation \\ 
\hline
\textit{Friends} & 
Home Exercise, Yoga, Gardening, Walking, Meditation \\ 
\hline
\textit{Public} & 
Posture Correction, Fall Risk Mitigation, Collision Alert, Gait Correction, Walking, Balance Tracking, Track Person, Health Diagnostics/Sensing, Picking up Items, Reminders, Fashion Use (Jewelry), Water Consumption Tracker, Feedback on Self Odor \\ 
\hline
\end{tabular}
\end{table*}

\begin{table*}[ht!]
\centering
\caption{Codebook used during thematic coding}
\begin{tabular}{|p{0.95\textwidth}|}
\hline
\textbf{Category: Location} \\
any where, beach, bedroom, car, care facility, exercise class, garden, gym, home, hospice, indoors, outdoors, PT office, shopping, train, work \\ \hline
\textbf{Category: Co-Presence} \\
caregivers, children, class member, doctor, doctors, family, friend, instructor, no one, pets, physical therapist, sexual partner, significant other, technician \\ \hline
\textbf{Category: Human-Kinesiology} \\
high movement, low movement, sedentary\\ \hline
\textbf{Category: Human-Conditions} \\
Deaf, Blind, Cognitive \\ \hline
\textbf{Category: Robot-Role} \\
sentinel, companion, coach, tool \\ \hline
\textbf{Category: Robot-Usage Temporal} \\
frequency-daily, lifetime, timespan-longterm, when-always, when-desired, when-in need, when-in-therapy, when-on schedule \\ \hline
\textbf{Category: Robot-Embodiment} \\
affordances, clothing-site/application specific, concern-arachnophobia, concern-disability-indicator, concern-distracting usage in public, concern-embarrassing usage in public, concern-falling off, concern-finding-robot, concern-goofy, concern-how do I use this, concern-hurt, concern-interference with handbags/grocery bags, concern-noisy-operation, concern-overreliance, concern-pet-friendliness, concern-privacy, concern-sensory-overload, concern-sitting-on, concern-social-isolation, concern-taking-place-of-humans, concern-trendy usage in public, concern-weird usage in public, embodiment-fitted-to-body, embodiment-less-visible, easy-to-clean, easy-to-remove, easy-to-use, form-factor-large, form-factor-sturdy, number-swarm, sensory-non-obtrusive, sensory-soft, sensory-voice, sensory-weight, sensory-smell, visuals-abstract, visuals-bright, visuals-customization, visuals-cute, visuals-less visible, visuals-utilitarian, embodiment-cute, voice-cute \\ \hline
\textbf{Category: Robot-Topology} \\
ankle, arm, back, belly, chest, hand, head, hip, knee, leg, neck, on body, shoulder, thighs, wrist \\ \hline
\textbf{Category: Robot-Perception} \\
environment-branches, environment-fruit-ripeness, environment-hazards, environment-humidity, environment-sound, environment-temperature, environment-ontrack, environment-visibility, environment-weather, human-muscle-tension, human-actions, human-balance, human-blood-pressure, human-blood-sugar, human-body-position, human-body-temperature, human-comfort, human-dizzy, human-EMG, human-form, human-goal-tracking, human-happiness, human-heartrate, human-intention, human-muscle-activity, human-nerves, human-nervousness, human-O2, human-odor, human-pain, human-range-of-motion, human-ready-state, human-sound, human-sweat, human-tired, human-pressure-applied, self-movement, vision \\ \hline
\textbf{Category: Robot-Autonomy} \\
determining plan, incorporating feedback, adjusting plan, full user control, intelligence \\ \hline
\textbf{Category: H2R Communication Modality} \\
button, gesture, physical movement, remote control, slap, smartphone app, sound, squeeze, taps, verbal, vibration, visuals \\ \hline
\textbf{Category: R2H Communication Modality} \\
blinks, cold, heat, kinesthetic, non-verbal sounds, physical movement - location, physical movement - patterns, vibrotactile-buzz, verbal, visual -- colors \\ \hline
\textbf{Category: Communication Frequency} \\
H2R-instant, R2H-instant, R2H-cyclic-feedback-requesting, R2H-cyclic-feedback-giving, R2H-cyclic \\ \hline
\textbf{Category: Communication Content} \\
H2R-corrective, H2R-informative, H2R-reactive, R2H-corrective, R2H-informative, R2H-informative dangers, R2H-informative feedback session, R2H-informative feedback explanatory, R2H-informative post usage, R2H-informative shutdown, R2H-informative to other people, R2H-reactive, R2H-supportive \\ \hline
\textbf{Category: Applications} \\
Acupressure/Acupuncture, Massage, Balance Monitoring, Body Hygiene Monitor, Chewing Support, Collision Alert, Dance, Fall Risk Mitigation, Walking, Fashion Use (Jewelry), Gait Correction, Gardening, Grocery Shopping Helper, Health diagnostics/sensing, Heart activity Monitoring, SOS alert, Navigation, Home Exercise, Meditation, Monitoring by caregivers, Physical Therapy, Picking up Items, Pleasure, Posture Correction, Reminders, Health data monitoring by doctors, Sleep Aid, Step tracks, Sugar levels Monitoring, Topical Medicine Application, User Motivator (through challenges), Water Consumption Tracker, Yoga \\ \hline
\textbf{Category: Miscellaneous} \\
Charger, On-boarding/Instructions, Difficulty in learning New Technology, Economic concerns, Ease of Use, Replacing Humans, Social Isolation, Personalization, Proactive Information Giving, Crawling, Social Interaction, Social Design Factors, Implicit sensing \\ \hline
\end{tabular}
\end{table*}